\documentclass{article}
\usepackage{ijcai19}

\usepackage{times}
\usepackage{soul}
\usepackage{url}
\usepackage[hidelinks]{hyperref}
\usepackage[utf8]{inputenc}
\usepackage[small]{caption}
\usepackage{graphicx}
\usepackage{amsmath}
\usepackage{booktabs}
\usepackage{algorithm}
\usepackage{algorithmic}
\usepackage{float} 
\usepackage{subfigure}
\usepackage{xspace}
\usepackage{amsthm}

\newcommand{\sys}{Combo\xspace}
\newcommand{\eg}{e.g.,\xspace}

\title{Decentralized Federated Learning: A Segmented Gossip Approach}

\author{
Chenghao Hu$^1$\and
Jingyan Jiang$^2$\and
Zhi Wang$^{3}$\footnote{Corresponding Author}\\
\affiliations
$^1$Department of Computer Science and Technology, Tsinghua University\\
$^2$College of Computer Science and Technology, Jilin University  \\
$^3$Graduate School at Shenzhen, Tsinghua University \\
$^3$Peng Cheng Laboratory\\
\emails
huch16@mails.tsinghua.edu.cn,
jiangjy14@mails.jlu.edu.cn,
wangzhi@sz.tsinghua.edu.cn
}

\begin{document}

\maketitle

\begin{abstract}

The emerging concern about data privacy and security has motivated the proposal of \emph{federated learning}, which allows nodes to only synchronize the locally-trained models instead their own original data. Conventional federated learning architecture, inherited from the parameter server design, relies on highly centralized topologies and the assumption of large nodes-to-server bandwidths. However, in real-world federated learning scenarios the network capacities between nodes are highly uniformly distributed and smaller than that in a datacenter. It is of great challenges for conventional federated learning approaches to efficiently utilize network capacities between nodes. In this paper, we propose a model segment level decentralized federated learning to tackle this problem. In particular, we propose a segmented gossip approach, which not only makes full utilization of node-to-node bandwidth, but also has good training convergence. The experimental results show that even the training time can be highly reduced as compared to centralized federated learning.

\end{abstract}

\section{Introduction}


Recent years have witnessed a rapid growth of deep learning algorithms which achieve and even transcend the human-level accuracy on nature language processing and computer vision \cite{devlin2018bert:,he2016deep}, thanks to the massive amount of data collected. To improve the deep learning performance, it is of great demand for different entities to contribute their own data and train models together. In such collaborative training, the concern about data leakage has motivated \emph{federated learning} \cite{McMahan2017FL}, which allows nodes to only synchronize the locally-trained models instead of their own original data. 

A general federated learning system uses a central parameter server to coordinate the large federation of the participating workers (workers and nodes are used interchangeably in this paper). The workers train a local model with their own dataset and send the model updates (\eg gradients or parameters) periodically to a centralized server for synchronization. To reduce the risk of single point failure, a couple of decentralized synchronization methods have been proposed. All-reduce \cite{patarasuk2009allreduce} adopts an all-to-all scheme, i.e., each worker sends the local model updates to all other workers. It achieves the same synchronization effect as parameter server but consumes much bandwidth resource between works. When the model updates from all nodes in the system are sent to all other nodes, the performance is highly degraded. To reduce the transmission cost, gossip based model synchronization \cite{daily2018gossipgrad:,haas2002gossip-based} is proposed: workers send local updates to only one or a group of selected nodes.

In real-world federated learning scenarios, the network capacities between nodes are highly uniformly distributed and smaller than that in a datacenter \cite{vulimiri2015global}. Thus, it is still extremely bandwidth costly when workers send the \emph{full} model updates (e.g., the size can be up to $1360$MB in $BERT_{LARGE}$ \cite{devlin2018bert:}). An intuitive question is then, is it possible for workers to synchronize the model \emph{partially}, from/to \emph{only} a part of the workers, and still achieve good training results? 

Our answer to this question is a novel decentralized federated learning design, introducing a segmented gossip approach, which not only makes full utilization of node-to-node bandwidth by transmitting model \emph{segmentations} in a peer-to-peer manner, but also has good training convergence by carefully forming dynamical synchronization gossiping groups. In particular, the details of the design and the contributions are summarized as follows.

First, we propose a model segmentation level synchronization mechanism. We ``split'' a model into a set of segmentations---subsets which contain the same number of model parameters that are not overlapped with each other. Workers perform segmentation level update by aggregating a local segmentation with the corresponding segmentation from $k$ other workers. Based on our analysis, $k$ can be much smaller than the number of all workers, to achieve good convergence for the training process.

Second, we propose a decentralized federated learning design, borrowing the idea from gossip protocol; each worker stochastically selects a few workers to transfer the model segment for each training iteration. Our objective is to maximize the utilization of bandwidth capacities between workers. To improve the convergence performance of our solution, we introduce ``Model Replica'' to guarantee that enough information from different workers is acquired during aggregation.

Third, we implement the model segmentation strategy and the gossiping strategy into a prototype called \sys, and design experiments to evaluate its performance. Our results show that our design significantly reduces the training time in practical network topology and bandwidth setup, with only slight accuracy degradation.

\section{Related work}
\textbf{Distributed ML}
Conventional distributed machine learning systems are centralized, workers periodically send the local updates to a (a set of) parameter servers (PS), such as SparkNet \cite{moritz2016sparknet:}, Tensorflow \cite{abadi2016tensorflow:} and traditional federated learning systems \cite{konecny2015federated,konevcny2016federated}. To avoid bottleneck and single point failure, \cite{li2014scaling,mlsys2019towards} aim to scale PS for better network utilization. Although these scaling methods could increase the accumulative bandwidth at the server side, they are still suffering the long convergence time when the network is poor.

An alternative solution is decentralized architecture, the workers exchange updates directly using all-reduce scheme, with a communication cost $O(n^2)$ for $n$ workers. To reduce the huge communication costs, an intuitive approach is to take the advantage of topology. Baidu first introduced Ring-allreduce\footnote{\url{http://andrew.gibiansky.com/blog/machine-learning/baidu-allreduce/}}, which is a bandwidth-optimal way to do an allreduce. The workers involved are arranged in a ring, each worker sends gradients to the next clockwise worker and receives from the previous one. In this way, it reduces the communication complexity to linear growth in scale. similarly, the tree \cite{li2015malt:} and graph \cite{agarwal2014a} topologies are proposed to reduce the communication cost. However, these approaches may need multiple hops between workers, resulting in slow convergence. Instead of the topology-based method, Ako \cite{watcharapichat2016ako:} propose a partial gradient updates method. In each synchronization round, each worker sends a gradient partition to every other worker. Obviously, Ako reduces the synchronization time and the communication overhead depends on the partition size and the worker number. 
 
Although these existing approaches perform well in distributed ML, they aggregate gradients every epoch, which still face heavy communication cost and is not practical in federated learning with slow internet connections.

\textbf{Communication efficient FL} 
A main research focus of federated learning is to reduce the communication cost. \cite{konevcny2016federated} propose structured updates and sketched updates to reduce the exchange data size at the cost of accuracy loss. \cite{McMahan2017FL} propose the federated averaging algorithm (FedAvg) to reduce the parameter updates significantly. FedAvg aggregates parameters after several epochs. In each synchronization round, it selects a fraction of workers and computes the gradient of the loss over all the data held by these workers. These methods are based on the PS architecture, which faces the network congestion when the updates arrive at the PS concurrently.

\textbf{Gossip protocol in ML}
The gossip protocol widely used in distributed systems \cite{baraglia2013a,haas2002gossip-based}, each worker sends out message to a set of other workers, the message propagates through the whole network worker by worker. \cite{blot2016gossip} first introduced the gossip protocol in deep learning. They propose GoSGD, using sum-weight gossip protocol to share the updates with selective workers. The results show good consensus convergence properties. \cite{daily2018gossipgrad:} propose GossipGraD, which is a gossip based SGD algorithm for large scale deep learning system and reduces the communication complexity to $O(1)$. 

However, in federated learning, network connections between geo-distributed workers usually could not be fully utilized because of the bottleneck, which is ignored in these approaches.

\section{Segmented Gossip Aggregation}

Now consider the network topology with $n$ workers. An all-reduce worker pushes $n-1$ local model replicates to the other workers through $n-1$ links while a gossip worker is expected to push one local model replicate out through only one link. Within a datacenter where the workers are connected by the local area network, they can always communicate with each other at maximum bandwidth thus the gossip worker can achieve great speed up as the transmission size is drastically reduced.

However, in the federated learning context where the workers are geo-distributed, the real bandwidth between the workers is typically small due to the potential bottleneck of WAN. Thus the traditional gossip-based schemes can not make full use of the worker's bandwidth because the transmissions are limited in one or few links. We propose the \emph{Segmented Gossip Aggregation} to solve this problem by ``splitting" the transmission task and feeding them into more links.

\subsubsection{Segmented Pulling}

\begin{figure}[H]
\centering 
\subfigure[Segmented Pulling]{
\label{Fig: Segment sub.1}
\includegraphics[width=0.2\textwidth]{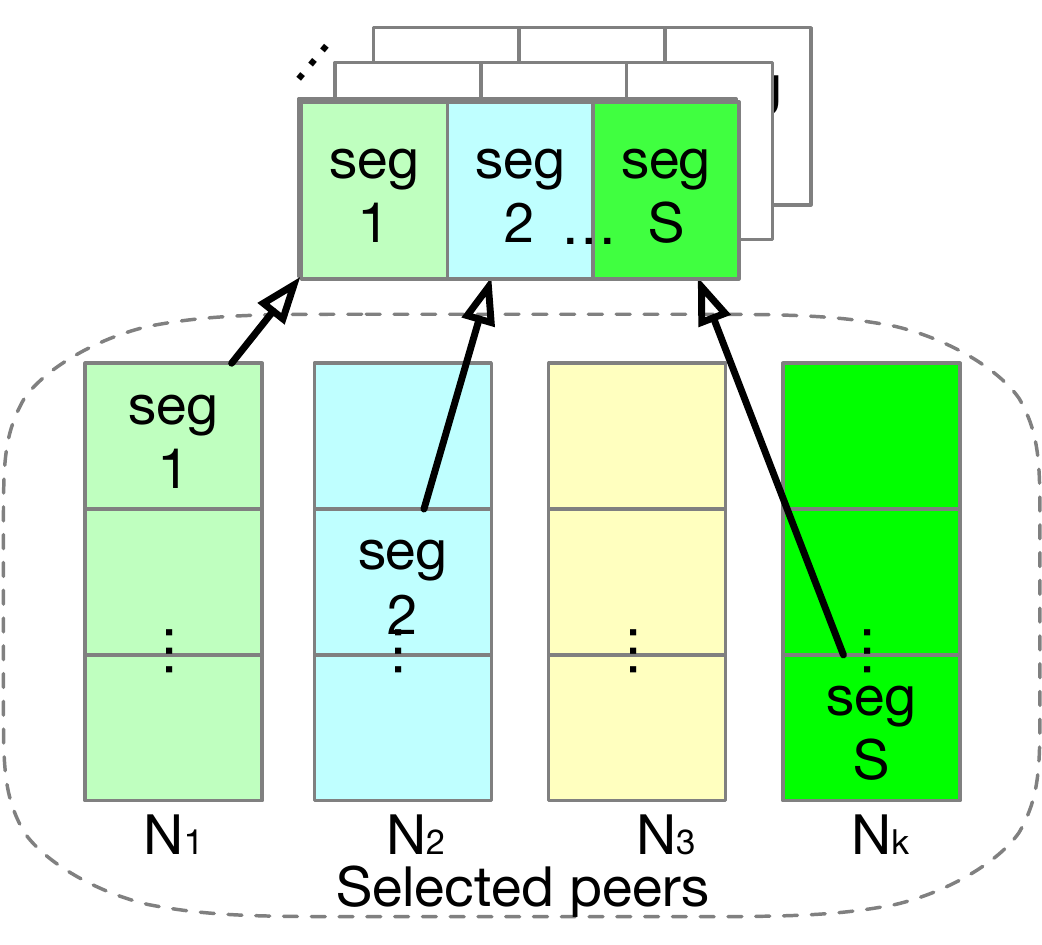}}
\subfigure[Segmented Aggregation]{
\label{Fig: Segment sub.2}
\includegraphics[width=0.225\textwidth]{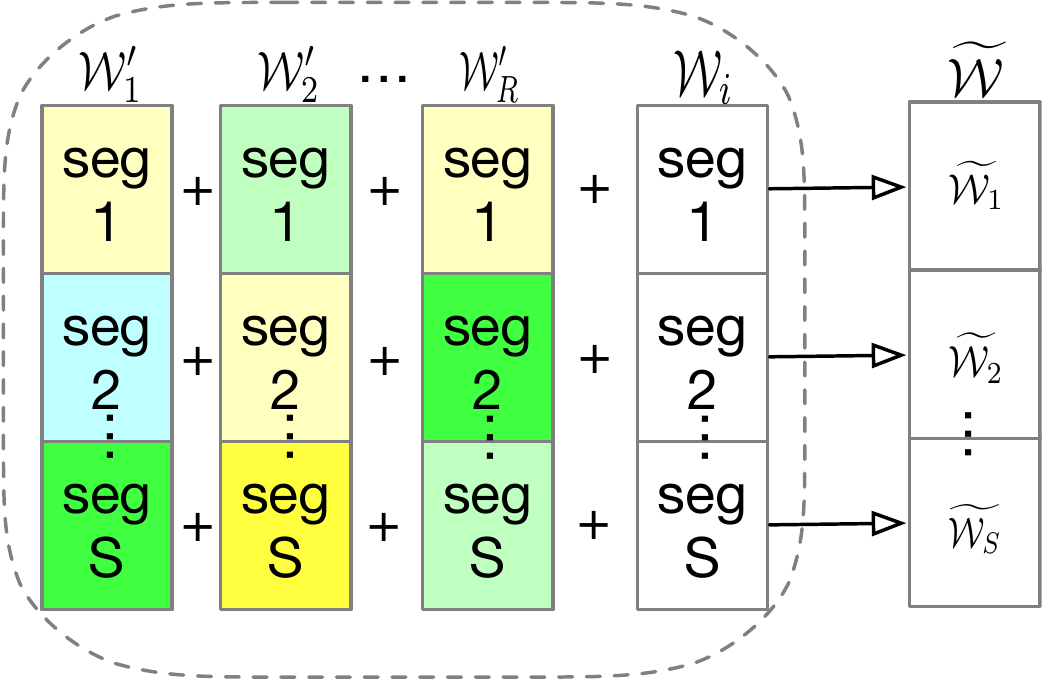}}
\caption{Segmented Gossip Aggregation}
\label{Fig: Segmented schema}
\end{figure}

Fig. \ref{Fig: Segment sub.1} illustrates the transfer procedure with segmented gossip aggregation which we name it \emph{segmented pulling.} In the aggregation phase, the worker needs to receive the model parameters from others. While the naive gossip-based synchronization schemes require the worker to collect the whole model parameters, segmented pulling allows the worker to pull different parts of the model parameters from different workers and rebuild a mixed model for aggregation.

Let $\mathcal{W}$ denote the model parameters. The worker firstly breaks the structure of $\mathcal{W}$ into $S$ segments without overlapping such that
\begin{equation}
    \mathcal{W} = (\mathcal{W}[1],\mathcal{W}[2],\dots,\mathcal{W}[S])
\end{equation}

For each segment $l$, the worker chooses a peer worker which we denote it as $j_l$ and then actively pulls the corresponding segment $\mathcal{W}_{j_l}[l]$ from it. Note that this step is parallelized to make full use of the bandwidth. When the worker fetches all the model segments back, a new mixed model $\mathcal{W}^\prime$ can be rebuilt from the segments such that
\begin{equation}
    \mathcal{W}^\prime = (\mathcal{W}_{j_1}[1],\mathcal{W}_{j_2}[2],\dots,\mathcal{W}_{j_l}[S])
\end{equation}

The naive gossip-based scheme pulls all the segments from a single peer worker. However, with segmented pulling, if we choose a different peer for each segment, the total transmission size is still equal to one model, like the naive gossip-based schemes, but the traffic is dissolved among not one but $S$ links.

\subsubsection{Model Replica}

In traditional distributed ML scenario within the datacenter, the gossip-based solutions can choose only one other worker for aggregation but still achieve excellent convergence, because the workers ``gossip" with each other frequently such that the update of each worker are propagated through the whole network before they become too stale \cite{daily2018gossipgrad:}. However, for communication efficient FL systems, the staleness of the model updates is hard to bound as the models are trained separately for up to a few epochs.

Thus as a compromise, we set a hyper-parameter \textit{Model Replica} $R$ which represents the number of the mixed model gathered by segmented pulling. To rebuild $R$ mixed models, the worker will pull $S \times R$ segments from peers. Thus increasing the value of $R$ means more segments have to be transferred through the network, which may cause bandwidth overhead. But this is necessary to accelerate the propagation and ensure the model quality. Since there is no centralized server bottleneck, the model training speed could still be faster even with extra transmission.

\subsubsection{Segmented Aggregation}

Typically the model aggregation uses weighted averaging of the received model parameters with the worker's dataset size as weight. But in segmented gossip aggregation, the mixed models are patched together from different workers, so it is hard to set a reasonable weight for the mixed model as a whole. For such case, we use a segment-wise model aggregation. 

 Assume the worker $i$ has fetched all the segments and rebuilt $R$ mixed models which we represent as $\mathcal{W}^\prime_1,\mathcal{W}^\prime_2, \dots ,\mathcal{W}^\prime_R$. Then for each segment $l$, we have $R$ mixed models and one local model to aggregate. Let $P_l$ denote the set of the workers which provide the segment $l$ (worker $i$ itself is contained too) and $|D_j|$ denote the dataset size of worker $j$, then we can aggregate segment $l$ by:
 
 \begin{equation}
 \label{eq:seg_agg}
     \widetilde{\mathcal{W}}[l] = \frac{\sum_{j\in P_l}|D_j|\mathcal{W}_j[l]}{\sum_{j\in P_l}|D_j|}
 \end{equation}

Combine all the aggregated segments, and we can rebuild the final aggregation result by 
\begin{equation}
    \mathcal{W} = (\widetilde{\mathcal{W}}[1],\widetilde{\mathcal{W}}[2],\dots,\widetilde{\mathcal{W}}[S])
\end{equation}
And then the worker can continue its training until next aggregation phase comes.

\section{\sys Design}
In this section, we introduce \sys, a decentralized federated learning system based on segmented gossip aggregation. We firstly present the implementation details of \sys, then discuss how it handles the dynamic nature of FL workers, and finally, we give a brief analysis of the convergence of \sys.

\subsection{Implementation Details}
As a decentralized FL system, we focus on the design of the workers as the participation of the centralized server is trivial during the training. However, it is important to notice that before the training starts, the server has to initialize the model parameters of each worker with the same value otherwise the training may fail to converge. The server has the information of all the workers, and while initializing the parameters, the worker list is also broadcasted.

\begin{figure}[H]
\centering 
\includegraphics[width=0.49\textwidth]{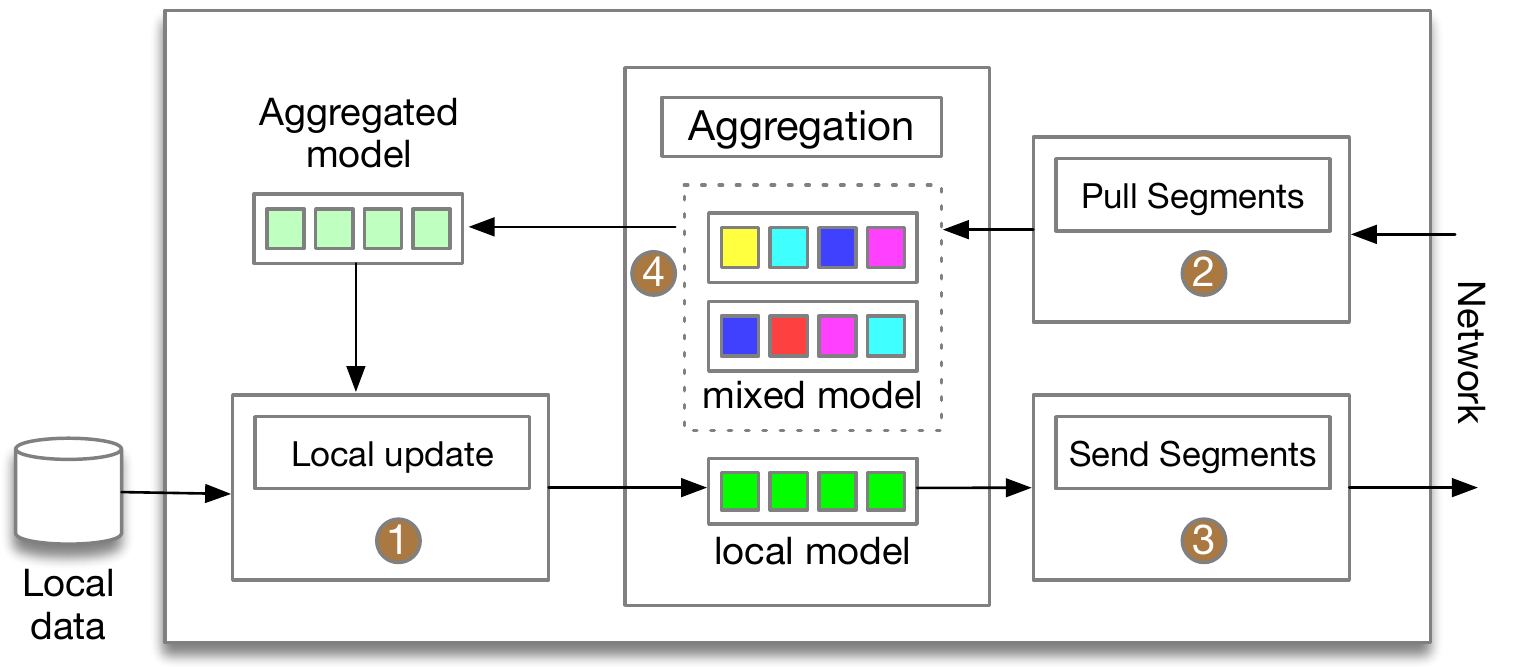}
\caption{The architecture of \sys workers}
\label{Fig: worker-arch}
\end{figure}

A \sys worker follows a stateful training process as illustrated by the numbered steps in Fig.\ref{Fig: worker-arch}. At each iteration, the workers (1) update the model with local dataset and meanwhile, (2) send the segment pulling requests to other workers, once the update is finished, they (3) send the segments to the requestors as a response of the pulling requests and when all the pulling requests are satisfied, the workers (4) aggregate the model segments and start next iteration. Next, we describe the implementation details of these steps.

\noindent{\bf (1) Local Update.} The learning process starts with the worker updating the model with the local dataset. The worker takes the aggregation result of the last iteration as the input model and updates it using stochastic gradient descent(SGD) with the local data. To reduce the communication cost, the local update may contain multiple SGD rounds before the communication with other workers. We denote the communication interval or the number of SGD rounds as $\tau$, which, in typical FL systems, could be up to a few epochs.

\noindent{\bf (2) Segments Pulling.} The workers firstly decide how to partition the model. They don't have to follow the same partition rule, but for simplicity, we assume they partition the model into $S$ segments in the same way. For each segment, the worker has to select $R$ peers and sends the \emph{pulling request}, which contains a segment description and a unique identifier of the worker to indicate which part of the model is to be sent and whom it suppose to be sent to. 

Each worker has to send $S\times R$ segment pulling requests to the other workers, and \sys tries to distribute these requests evenly among all the workers to engage more links and balance the transmission workload. Thus for each request, the target worker is randomly selected from all the other workers without replacement until there is no option left, which means when $S\times R \leq n$, all the segments come from different workers. Note that for each iteration, the pulling requests can be sent even before the local update starts; in this way, the target workers can send the segments immediately when the local model is ready.

\noindent{\bf (3) Segments Sending.} The sending procedure is a twin action of the segments pulling. When the worker finishes the local update, it is ready to send its update result to others. Rather than actively pushing the model, the worker only dispatches the model segments according to the received pulling requests.

\noindent{\bf (4) Model Aggregation.} While the worker is providing the model segments to others, it is also receiving the segments it has requested previously. The model aggregation phase is blocked until all the pulling requests are satisfied, then the worker aggregates the external model segments with the local model using \eqref{eq:seg_agg} and put the aggregated segments together to rebuild the model. With the aggregation result, the worker gets back to the first step and starts the next training iteration.

\subsection{Dynamic Workers}
In the context of federated learning, the participating workers are more likely to be mobile phones and embedded devices, which are often not connected to a power supply and stable network. Thus the workers in FL system are highly dynamic and unstable, and they can join and exit the federation at any time. 

Traditional distributed systems adopt the heartbeat packet and time threshold to check the status of the workers. However, these methods are not applicable with the FL system for the following two reasons: 1) The server has to maintain the heartbeat connection with all the participating workers which limits the scalability of the system. 2) The computation times of each worker vary significantly due to the difference in the computing devices and network environment.

Fortunately, the design of \sys allows us to solve this problem decently. If the worker exits accidentally, the pulling requests it sends to other workers can be canceled immediately when the target workers find it unreachable. For those workers who have requested segments from the offline worker, they can monitor the status of the target workers, and once they see the connection with the target worker is lost, they can mark it as offline, resend the request to another worker and stop pulling from the offline worker. If it is a false report due to the network fluctuation or the offline worker comes back, the offline flag can be removed as long as the communication is reestablished. 

The participation of a new worker is relatively easy to handle. When a new worker comes to the federation, it first registers itself on the server and requests a worker list. Then it pulls the segments and aggregates them as normal only without its local model. With the aggregation result, it can start the training with its local dataset. When it sends the pulling requests to the target workers, the target worker adds the newcomer to the worker list. Since the new worker sends the pulling requests to many workers in a single iteration, its existence will be quickly noticed by all other workers.

\subsection{Convergence Analysis}

\newtheorem{theorem}{\bf Theorem}
\newtheorem{define}{\bf Definition}
\newtheorem{assumption}{\bf Assumption}

Generally, the deep learning uses the gradient descent algorithms to find the model parameters that minimize a user-defined loss function which we denote it as $F(\mathcal{W})$. For the loss function, we make the following assumptions.

\begin{assumption}
{\bf (Loss function)} $F(\mathcal{W})$ is a convex function with bounded second derivative such that
\begin{equation}
    \mu \leq ||\nabla^2F(\mathcal{W})|| \leq L
\end{equation}
\end{assumption}

In a centralized learning system, the model parameters are updated with the gradient $\nabla F(\mathcal{W})$ calculated from the whole dataset. However, with the federated settings, the worker $i$ updates the model with the gradient of a subset of data and we denote it as $\nabla F_i(\mathcal{W})$. To capture the divergence of these two gradients, we make the next definition.

\begin{define}
    {\bf (Gradient Divergence)} For any worker $i$ and model parameter $\mathcal{W}$, We define $\delta$ as the upper bound of the divergence between local and global gradients.
    \begin{equation}
      || \nabla F_i(\mathcal{W}) - \nabla F(\mathcal{W})|| \leq \delta
    \end{equation}
\end{define}

For a worker $i$ in our proposed system, at iteration $t$, the local model parameter $\mathcal{W}_{t,i}$ is an aggregation result of the local model and a few mixed models rebuilt from segments. As a contrast, we denote $\mathcal{W}_{t}$ as the aggregation result of all the nodes, which is the output of $FedAvg$ algorithm. Like the gradient divergence, we define aggregation divergence to measure the aggregation result.

\begin{define}
    {\bf (Aggregation Divergence)} For any worker $i$ at iteration $t$, we define $\rho$ as the upper bound of the divergence between partial and global aggregation.
    \begin{equation}
      || \mathcal{W}_{t,i} - \mathcal{W}_{t}|| \leq \rho
    \end{equation}
\end{define}

With the above assumption and definitions, we can present the convergence result of \sys. 

\begin{theorem}
\label{trm:converge}
    Let $\mathcal{W}^*$ denote the global optimum and $\mathcal{W}_0$ denote the initial model parameters, worker $i$ performs gradient descent on local dataset for $\tau$ times with learning rate $\alpha \leq \frac{1}{L}$ and then pull the segments to aggregate, the aggregation result is $\mathcal{W}_{t,i}$, the convergence upper bound of \sys is given by
    \begin{equation}
            ||\mathcal{W}_{t,i}-\mathcal{W}^*|| \leq \theta^{t\tau} ||\mathcal{W}_{0}-\mathcal{W}^*|| + (1-\theta^{t\tau})[\frac{\rho}{1-\theta^\tau} + \frac{\alpha \delta}{1-\theta}]
    \end{equation}
    where $\theta = 1-\alpha \mu$.
\end{theorem}

Due to the limitation of the space,  we will provide detailed proof in the extended version. Note that this bound is characteristic of stochastic gradient descent bounds that it converges to within a noise ball around the optimum rather than approaching it. The gap between the output and optimum comes from two parts: the gradient divergence $\delta$ and the aggregation divergence $\rho$. The gradient divergence is related to the data distribution of each worker, which is the inherent drawback of the FL system.

According to the above inequality, the influence of $\rho$ is exacerbated when the communication interval $\tau$ increases. The aggregation divergence can be ameliorated by aggregating more models from other workers. This explains why we set a hyper-parameter $R$ to control the model replicas received from others. If we let $R=n-1$, the worker aggregates all the external models and the model divergence decreases to zero. In this situation, \sys degrades to the all-reduce scheme and has the same training result as the centralized way. However, we argue that the value of $R$ could be much smaller but still maintains the training efficiency, which is then validated in the evaluation.

\section{Performance Evaluation}

\subsection{Setup}

We conduct simulation experiments to evaluate our design. The evaluation can be divided into two parts. First, the stateful and synchronous nature of \sys allows us to simulate the training process sequentially, while logically, the training result is the same as the parallelized way. The training traces of each worker are then recorded, which contains the validation accuracies, training iterations, and corresponding synchronization partners. Second, we simulate the network topology and feed it with the training traces to estimate the training time. The specific settings are listed as follows:

$\triangleright$ \textit{Training settings.} We train a CNN model on CIFAR-10 dataset to evaluate the training ability of \sys. The dataset consists of  50,000 images for training and 10,000 for validation. The training data are randomly distributed among the workers without overlapping, and the validation data are shared among every worker. The CNN model is adopted from \cite{McMahan2017FL}, which is considered to be suitable for CIFAR-10 dataset. 

The models are trained on each worker using SGD algorithm with the same hyper-parameters, that a learning rate of 0.1 and a batch size of 128. Notice that we adopt a large learning rate simply to accelerate the training speed and it doesn't affect the comparison results. The synchronization interval is set to 40, which means every worker perform SGD updates on the local model for 40 times before it communicates with others.

$\triangleright$ \textit{Network settings.} We simulate a fully connected network topology among the workers. The maximum bandwidth limit of each worker is set to be 100Mbps. Moreover, to simulate the bottleneck of WAN, we set 10Mbps as the available bandwidth between two workers.

$\triangleright$ \textit{Comparison settings.} We compare \sys with (1) traditional federated learning system with \emph{FedAvg} algorithm in which all the workers participate and the server is randomly selected from them, and (2) naive \emph{gossip} approach without segmentation. To make them comparable, they are all simulated within the same network topology.

The communication behavior of \sys is controlled by two parameters: \emph{model segments} as $S$ and \emph{model replica} as $R$. In our following experiments, we set $S=10$ and $R=2$ by default, that is in the synchronization phase, the model parameters are flattened and then divided into ten segments equally. For each segment, the worker requests two replicas from other workers. The gossip approach is the special case of \sys when $S=1$, and it shares the same value of $R$ with \sys.

$\triangleright$ \textit{Performance Metrics.} The learning performance is measured by the convergence speed. We record the top-1 validation accuracies of the aggregated models at each round and then align the accuracies to the corresponding times. The time is acquired from our network simulator where the local update time is referenced from the real machine time of training with a GTX 1080 Ti graphics card, and the communication time is calculated according to the bandwidth limitations.

\subsection{Experiment Result}

We first evaluate the convergence speed and scalability of \sys in comparison to the other approaches, then we explore the advantages and disadvantages of the design of model segments and replicas and how they affect the training performance.

\subsubsection{Convergence Speed}

\begin{figure*}[htbp]

\begin{minipage}[t]{0.32\textwidth}
\subfigure[Convergence with 30 workers]{
\label{Fig: time_acc_30}
\includegraphics[width=0.9\textwidth]{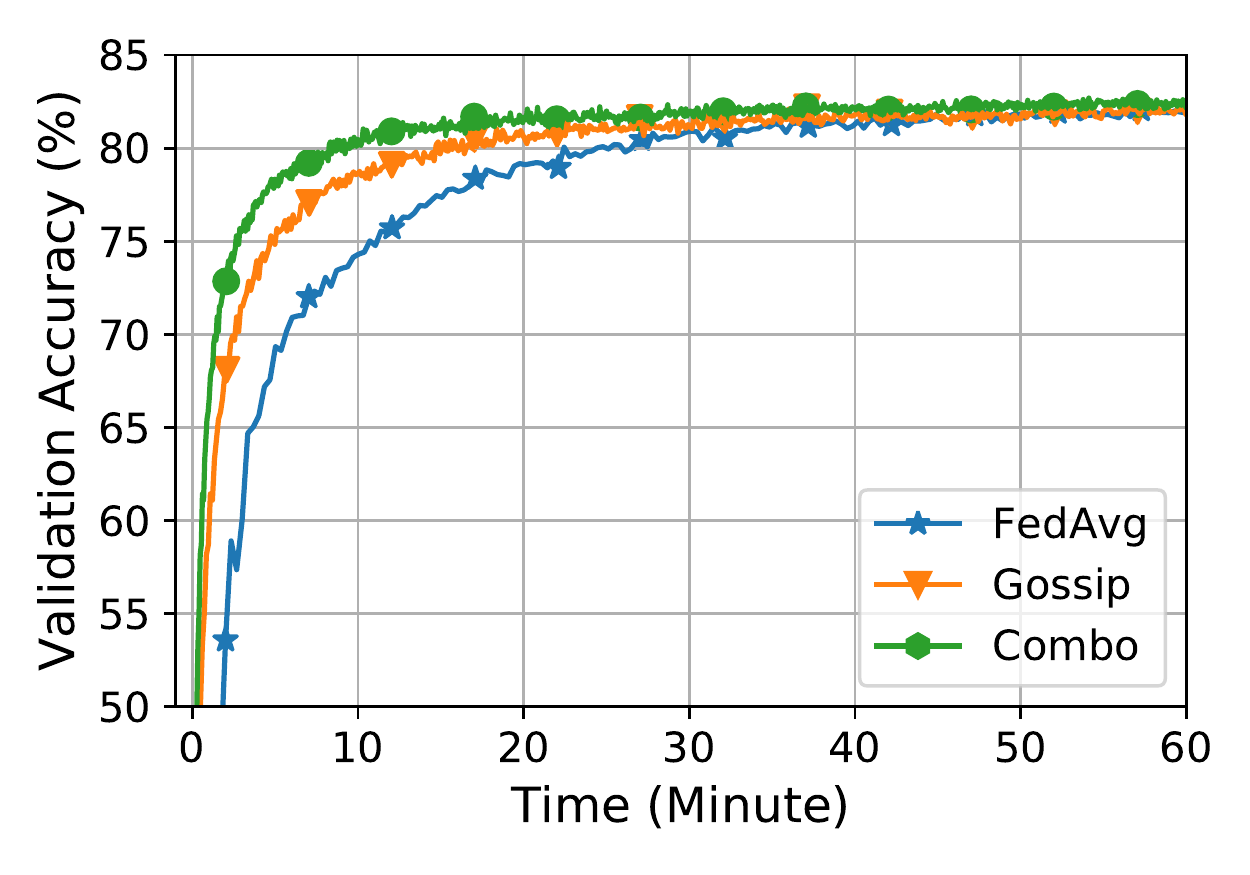}}
\\
\subfigure[Time to reach 80\% accuracy]{
\label{Fig: scale_time}
\includegraphics[width=0.9\textwidth]{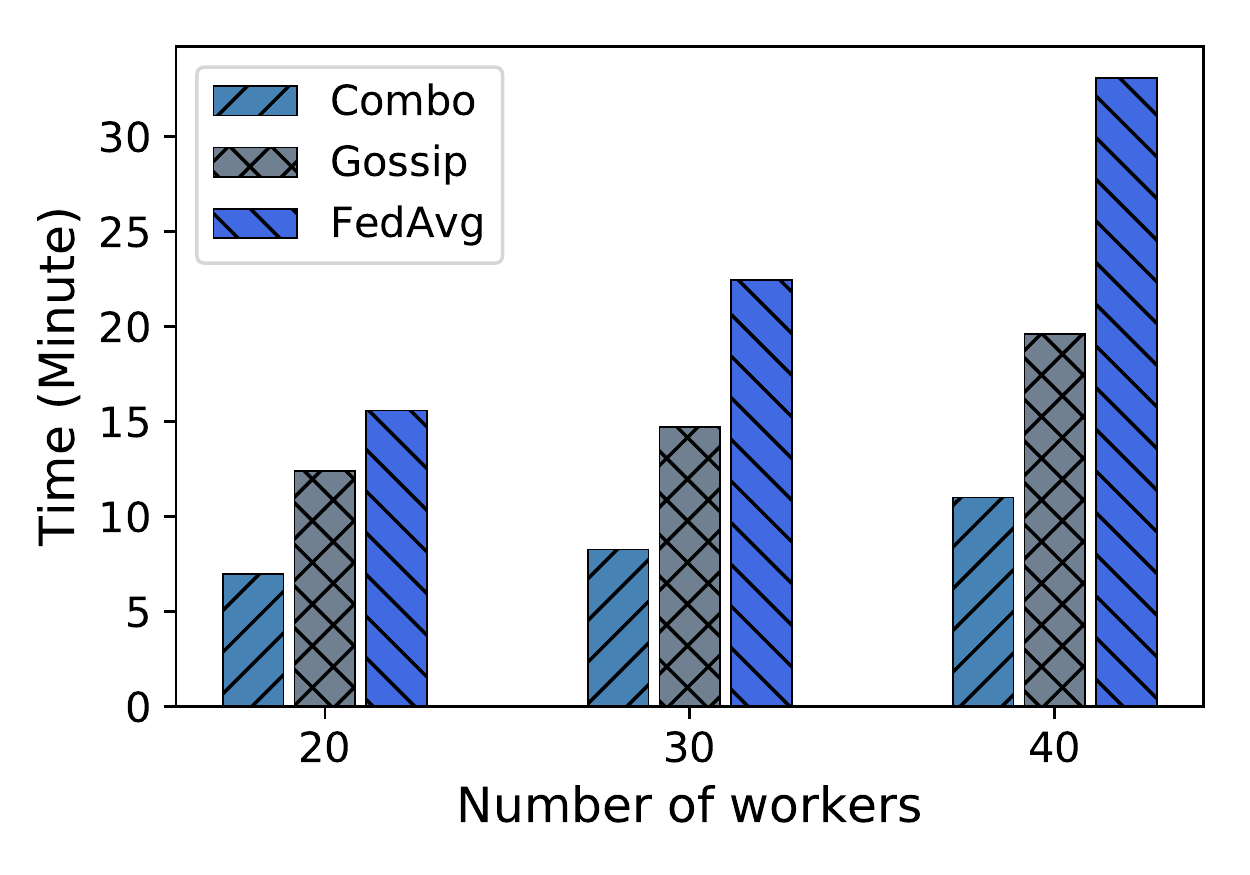}}

\caption{Convergence speed}
\label{Fig: acc}
\end{minipage}
\begin{minipage}[t]{0.32\linewidth}
\centering 
\subfigure[Convergence with model segments]{
\label{Fig: seg_acc}
\includegraphics[width=0.9\textwidth]{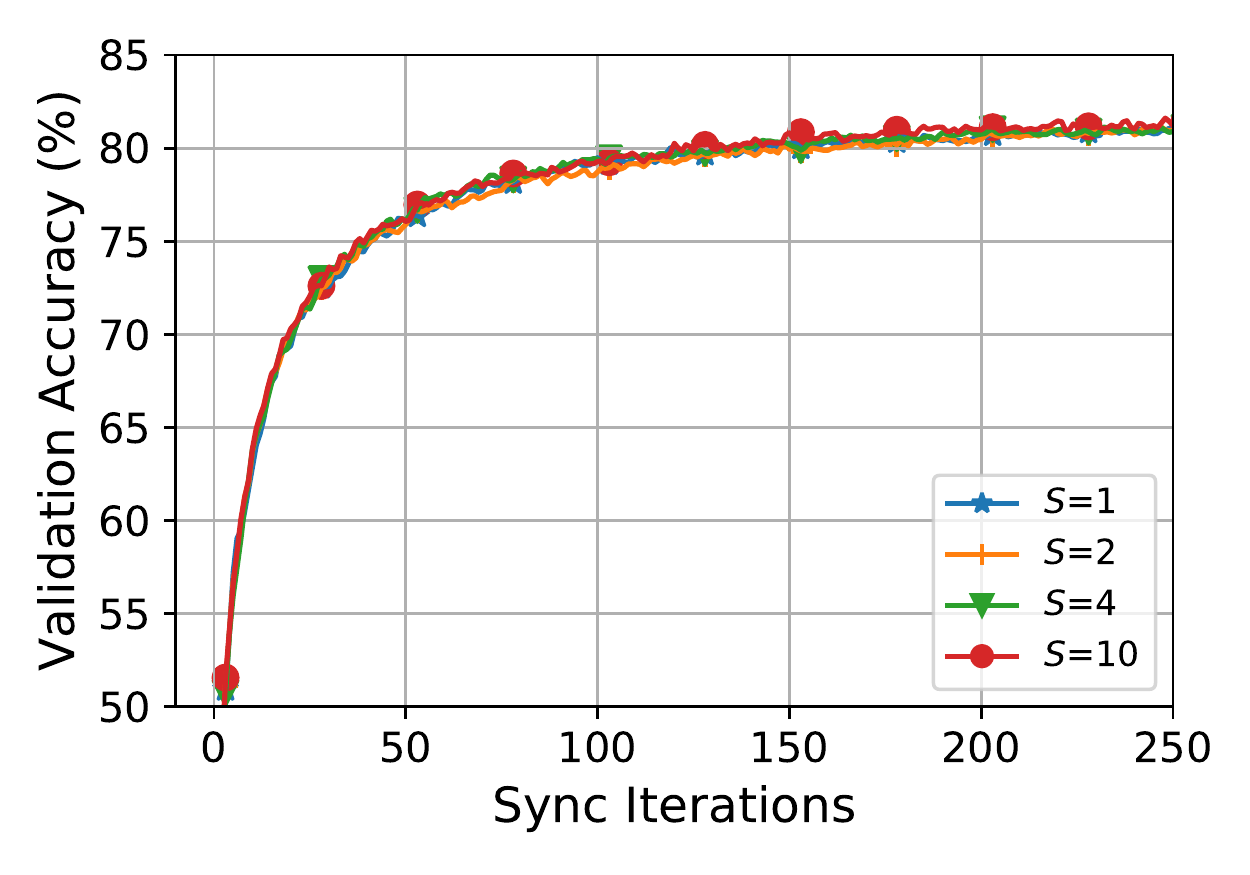}}
\\
\subfigure[Sync time comparison]{
\label{Fig: seg_time}
\includegraphics[width=0.9\textwidth]{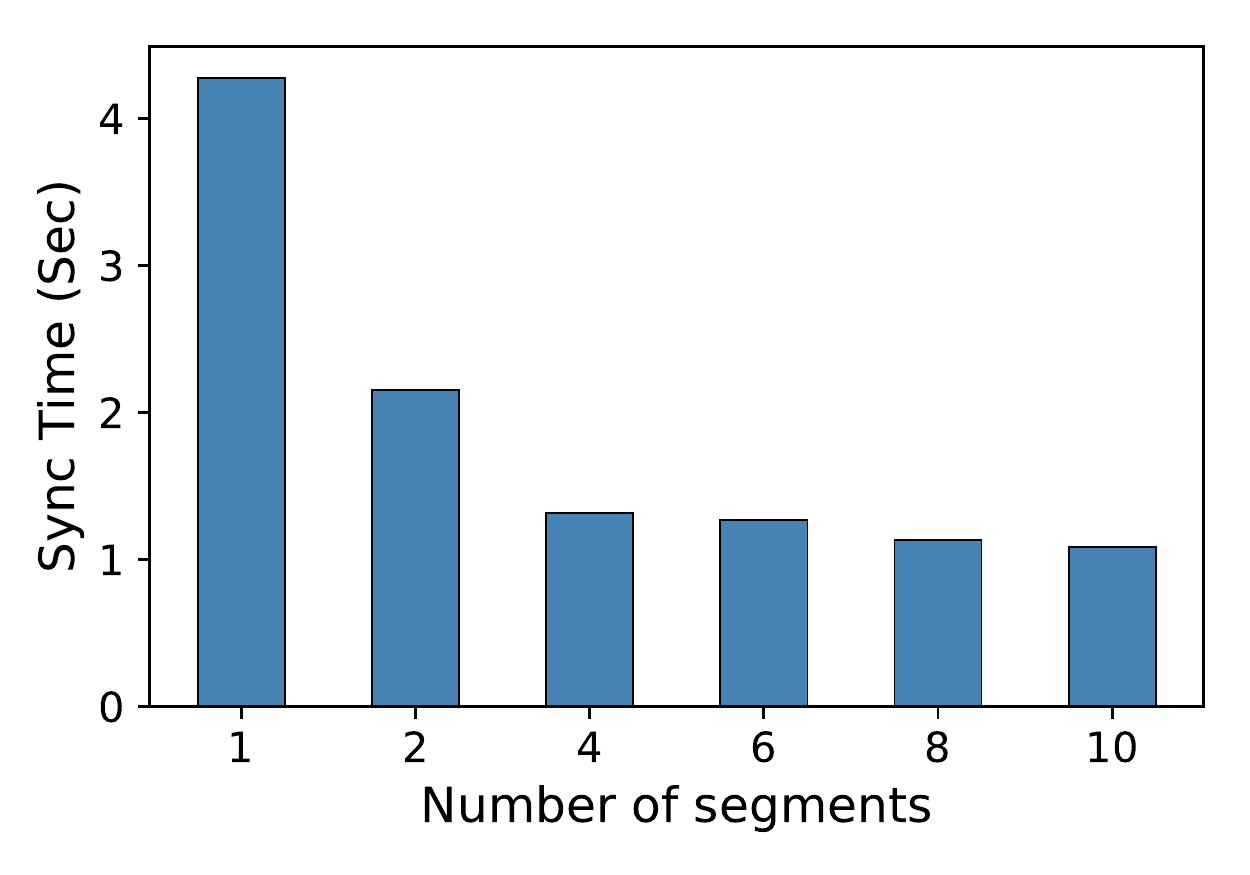}}
\caption{Benefit of model segments}
\label{Fig: acc}
\end{minipage}
\begin{minipage}[t]{0.32\linewidth}
\centering 
\subfigure[Convergence with model replicas]{
\label{Fig: replica_acc}
\includegraphics[width=0.9\textwidth]{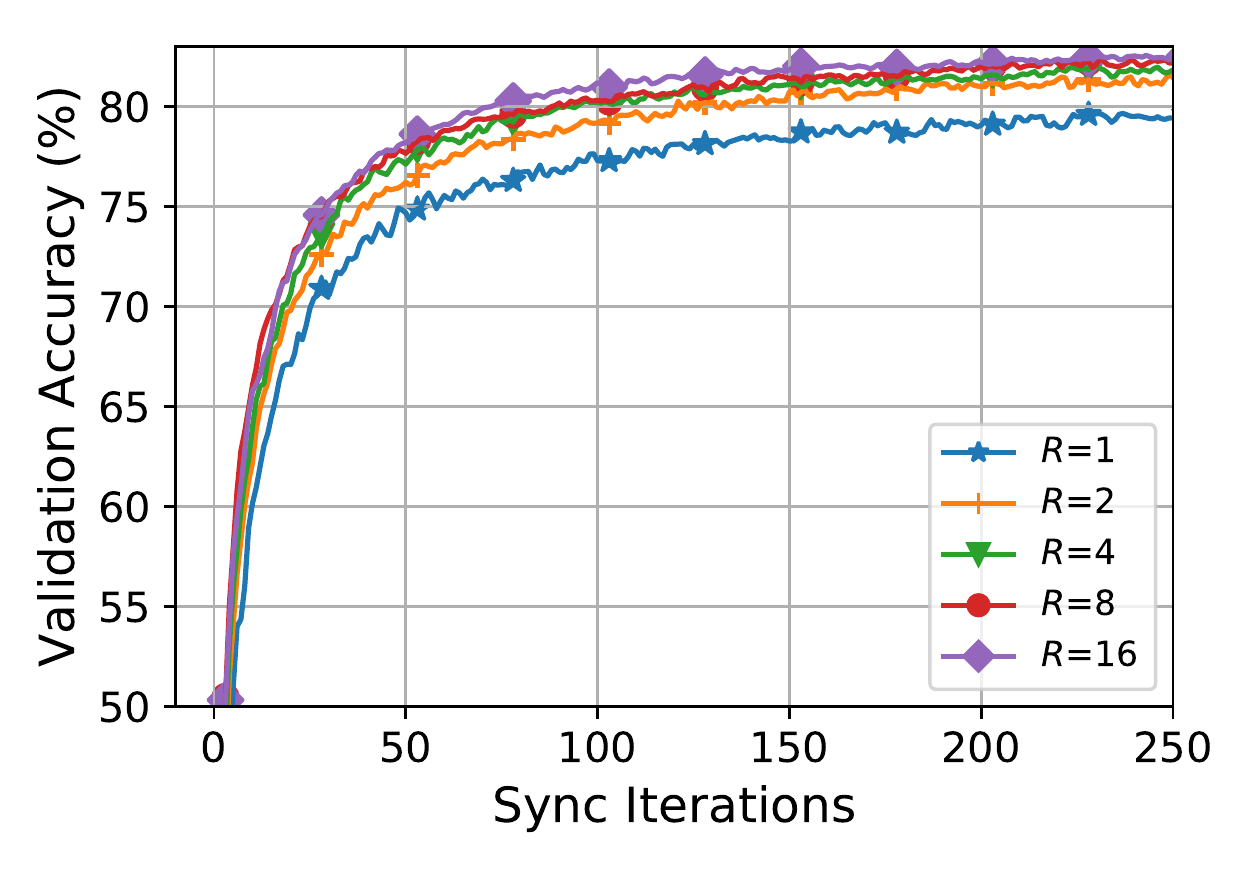}}
\\
\subfigure[Time to reach 80\% accuracy]{
\label{Fig: replica_time}
\includegraphics[width=0.9\textwidth]{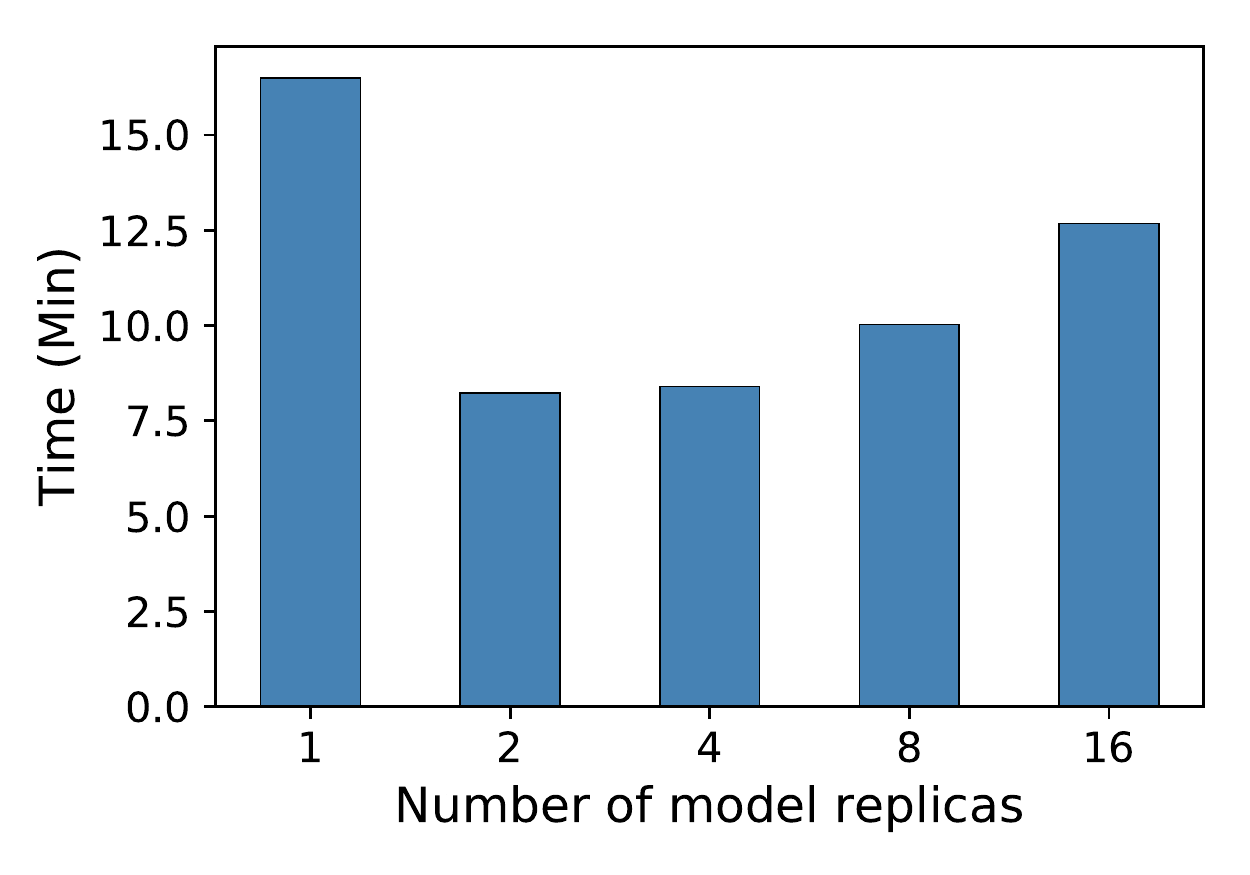}}

\caption{Impact of model replicas}
\label{Fig: acc}
\end{minipage}

\end{figure*}

We present the whole training process over time, as illustrated in Fig.\ref{Fig: time_acc_30}, \sys exhibits an apparent speedup in the convergence without affecting the final validation accuracy. We also explore the scalability of these three methods by comparing the training time needed to reach a predefined accuracy goal with varying number of workers among 20, 30, and 40. According to Fig.\ref{Fig: time_acc_30}, the model reaches convergence around 82\% validation accuracy. Since the aim is not achieving the best accuracy and practically speaking, it is not worthy of spending too much time for only 1\% or 2\% accuracy gain. Thus we set 80\% as the accuracy goal. 

As illustrated in Fig.\ref{Fig: scale_time}, \sys requires the least training time to reach the given accuracy within all three cases and compared with FedAvg algorithm, the speedup of \sys increases from $2.25\times$ to $3.01\times$ with the expansion of scale. This phenomenon indicates that the decentralized federated learning is more scalable than the centralized way within a peer-to-peer network.

\subsubsection{Benefit of Model Segments}
The speedup of decentralized approaches comes from the removal of the bottleneck of the centralized server, and the advantage of \sys comes from the benefit of model segments. We train the model with 30 workers, fix $R=2$ and vary $S$ from 1 to 10 to investigate how model segments affect the training performance.

Compared with the naive gossip solution, \sys aggregates mixed model parameters made up of multiple segments instead of the complete model. A potential concern is that the result may suffer degradation as the aggregation target is mottled and loses integrality. However, Fig.\ref{Fig: seg_acc} shows that the accuracy of the aggregated results at each synchronization iterations is not affected by the model segments at all. Partitioning the model into ten segments($S=10$) has the same convergence trend as that without partition.

While the model segments do not affect the accuracy at each iteration, the synchronization time is significantly reduced. As illustrated in Fig.\ref{Fig: seg_time}, by simply splitting the model parameters into two segments can reduce the synchronization time by half. This is because when $S=2$, the original transmission quantity is divided into two parts and fed into $2\times$ more links. When the bandwidth is not exhausted, the sending and receiving time can be reduced almost proportionally. However, when $S\geq 6$, the bandwidth is already fully exploited, increasing the number of segments will not improve the time consumption then.

\subsubsection{Impact of Model Replicas}

Next, we evaluate the impact of model replica, which controls the overall information quantity that the workers send and receive at each synchronization iterations. Similar to the previous settings, we fix $S=10$ and vary $R$ from 1 to 16.

As we discussed in the convergence analysis of \sys, the more information a worker receives, the better aggregation result it will get. When the worker receives all the model replicas from other peers, \sys becomes the All-reduce structure and achieves the same training result as the centralized approach. The analysis is validated by Fig.\ref{Fig: replica_acc} that with the increase of the number of model replicas, the accuracy of each iteration becomes better. However, the improvement is not unlimited. We can see that there is no significant gap between $R=8$ and $16$ in the convergence trend and result. This reflects the redundancy of All-reduce structure that the worker doesn't have to collect all the external models to train a high-quality model.

However, as the bandwidth of worker is fully utilized with model segments, increasing $R$ leads to the proportional growth of the transmission workload. Thus there exists a tradeoff, a larger $R$ increases the convergence rate on synchronization iterations but also the synchronization time. We compare the training time needed to reach 80\% validation accuracy with different $R$ as shown in Fig.\ref{Fig: replica_time}. Increasing $R=1$ to $2$ leads to a rapid reduction of the required training time as it drastically reduces the iterations needed to achieve target accuracy goal, which is also illustrated in Fig.\ref{Fig: replica_acc}. However, if we continue to increase $R$, the growth of the training time exceeds the reduction of the iterations and slows down the convergence speed.

\section{Conclusion}
One of the most challenging problem of federated learning is the poor network connection as the workers are geo-distributed and connected with slow WAN. To avoid the drawback of high possibility network congestion in centralized parameter sever architecture, which is adopted in today's FL systems, we explore the possibility of decentralized FL solution, called \sys. Taking the insight that the peer-to-peer bandwidth is much smaller than the worker's maximum network capacity, \sys could fully utilize the bandwidth by saturating the network with segmented gossip aggregation. The experiments show that \sys significantly reduces the training time and remains good convergence performance.

\section*{Acknowledgments}
This work is supported in part by NSFC under Grant No.~61872215 and 61531006, SZSTI under Grant No.~JCYJ20180306174057899, and Shenzhen Nanshan District Ling-Hang Team Grant under No.~LHTD20170005.

%
%
%
%
%
%

\bibliographystyle{named}
\bibliography{main}

\end{document}